\documentclass[runningheads]{llncs}
\usepackage{graphicx}
\usepackage{epstopdf}
\usepackage{amsfonts}
\usepackage{multirow}
\usepackage{hyperref}

\begin{document}
	\title{CETransformer: Casual Effect Estimation via Transformer Based Representation Learning }
	%
	%
	\author{Zhenyu Guo\inst{1,2}
		\and
		Shuai Zheng\inst{1,2}
		\and
		Zhizhe Liu\inst{1,2}
		\and
		Kun Yan\inst{1,2}
		\and
		Zhenfeng Zhu\thanks{Corresponding author}\inst{1,2}
	}

	\institute{Beijing Jiaotong University, Beijing, China\\
		\and
		Beijing Key Laboratory of Advanced Information Science and Network Technology, Beijing, China\\
		\email{\{zhyguo,zs1997,zhzliu,kunyan,zhfzhu\}@bjtu.edu.cn}
	}

	%
	%
	%
	\maketitle              
	\begin{abstract}
		Treatment effect estimation, which refers to the estimation of causal effects and aims to measure the strength of the causal relationship, is of great importance in many fields but is a challenging problem in practice. As present, data-driven causal effect estimation faces two main challenges, i.e., selection bias and the missing of counterfactual. To address these two issues, most of the existing approaches tend to reduce the selection bias by learning a balanced representation, and then to estimate the counterfactual through the representation. However, they heavily rely on the finely hand-crafted metric functions when learning balanced representations, which generally doesn't work well for the situations where the original distribution is complicated. In this paper, we propose a CETransformer model for casual effect estimation via transformer based representation learning. To learn the representation of covariates(features) robustly, a self-supervised transformer is proposed, by which the correlation between covariates can be well exploited through self-attention mechanism. In addition, an adversarial network is adopted to balance the distribution of the treated and control groups in the representation space. Experimental results on three real-world datasets demonstrate the advantages of the proposed CETransformer, compared with the state-of-the-art treatment effect estimation methods.
		\begin{center}
			
		\end{center}
	\end{abstract}

	\section{Introduction}
	Causal effect estimation is an crucial task that can benefit many domains including health care \cite{alaa2017bayesian,glass2013causal}, machine learning\cite{zhang2015multi,kuang2017estimating}, business\cite{wang2015robust} and sociology science\cite{gangl2010causal}. For example, in medicine, if two pharmaceutical companies have both developed anti-hyperlipidemic drugs, which one is more effective for a given patient? Suppose we consider different anti-hyperlipidemic drugs as different treatments, the therapeutic effects of the drugs can be obtained by estimating causal effects. As described above, the causal effect is used to measure the difference in outcomes under different interventions.
	
	In practice, we often obtain drug treatment effects by means of randomized controlled trials(RCT), and similar methods such as A/B tests are used to obtain average effects of a new feature in recommendation systems\cite{yin2019identification}. However, for individual causal effects, we cannot collect them by means of RCT because we lack counterfactual outcomes. Since counterfactuals are not directly available, causal effect estimation through massive data has become an important task in the era of big data and has been widely adopted\cite{yao2018representation,yao2019ace,schwab2020learning}. Nevertheless, data-driven causal effect estimation approaches face two main challenges, i.e., \textbf{treatment selection bias} and \textbf{missing counterfactuals outcomes}. 
	
	Firstly, in contrast to RCT, treatments in observational data are usually not randomly assigned. In the healthcare setting, physicians consider a range of factors when selecting treatment options, such as patient feedback on treatment, medical history, and patient health status. Due to the presence of selection bias, the treated population may differ significantly from the controlled population. Secondly, in real life, we only observe the factual outcome and never all the potential outcomes that could have occurred if we had chosen a different treatment option. However, the estimation of treatment effects requires to compare the results of a person under different treatments.  These two issues make it challenging to obtain an assessment of the treatment effect from the observed data.
	
	To address both of these challenges, existing approaches\cite{johansson2016learning,schwab2020learning} project the observed data into a balanced representation space where different treatment groups are close to each other, and then train an outcome prediction model to estimate the counterfactual. To the best of our knowledge, existing methods use finely hand-crafted metric functions to approximate the distribution of different treatment groups, and the network structures are the simplest fully connected neural networks. In \cite{yao2018representation}, the authors considered propensity scores as the relative position of individuals in the covariate space and performed the construction of triplet pairs. They adopted a hand-crafted metric function between the midpoint from different treatment groups  to balance the distribution. By adopting an integral probability metric(IPM)\cite{yao2019ace}, the similarity between two distributions is measured. The network architecture used in all of the above approaches is the most primitive fully connected network. To reduce the difference between different distributions, using manually designed similarity metric functions alone is not well adapted to the situation where the original distribution is complicated. Meanwhile, more attention should be paid to the correlation between covariates to generate a more discriminative representation, while the fully connected network is shown to exploit only the relationship between individuals\cite{domingos2020every}.
	
	Aiming at solving the above problems, we propose an casual effect estimation model via transformer based representation learning(CETransformer). The key contributions of this work are as follows:
	\begin{itemize}
		\item {The recently popular Transfomer network is adopted as a converter for mapping individuals into a latent representation space. Hence, the correlation between different covariates can be effectively exploited through self-attention mechanism to benefit the representation learning.}
		\item {To make the transformer trainable in the situation of limited number of individuals, 
			we take a form of self-supervision via auto-encoder to realize the augmentation of training data.}
		\item {Rather than simply using a metric function, the adversarial learning is utilized to balance the distribution between the different treatment groups. }
	\end{itemize} 
	We organize the rest of our paper as follows. Technical background including the basic notations, definitions, and assumptions are introduced in Section \ref{Preliminary}. Our proposed framework is presented in Section \ref{Methods}.  In Section \ref{Experiments}, experiments on three public datasets  are provided to demonstrate the effectiveness of our method. Finally, Section \ref{Conclusion} draws our conclusions on this work.
	
	\section{Preliminary \& Background}
	\label{Preliminary}
	Suppose that the observational data $X =\{X_i \in \mathbb{R}^d\}_{i=1}^n$ contain $n$ units(individual\\-s/samples) with each containing $d$ feature variables, and that each individual received one of two treatments. Let $T_i$ denote the binary treatment assignment on unit $X_i$, i.e., $T_i = 0$ or $1$. For the unit $X_i$ in the treated group, $T_i = 1$, and it will belong to the control group if $T_i = 0$. Before the treatment assignment, any outcome $Y_1^{i}$(treated) or $Y_0^{i}$(control), is taken as a \emph{potential outcome}. After the intervention, the outcome $Y_{T_i}^{i}$ will be the \emph{observed outcome} or \emph{factual outcome}, and the other treatment’s outcome $Y_{1-T_i}^{i}$ is the  \emph{counterfactual outcome}. 
	
	Throughout this paper, we follow the potential outcome framework for estimating treatment effects\cite{rubin1974estimating,splawa1990application}. Specifically, the individual treatment effect
	(ITE) for unit  $x_i$ is defined as the difference between the potential treated and control outcomes:
	\begin{equation}
		ITE_i = Y_1^i - Y_0^i, (i=1,\cdots,n).
	\end{equation}
	Meanwhile, the average treatment effect (ATE) is the difference between the potential treated and control outcomes, which is defined as:
	\begin{equation}
		ATE = \frac{1}{n}\sum_{i=1}^{n}(Y_1^i - Y_0^i), (i=1,\cdots,n).
	\end{equation}
	
	Under the potential outcome framework, the common assumptions to ensure the identification of ITE  include: \emph{Stable Unit Treatment Value Assumption (SUTVA)}, \emph{Consistency}, \emph{Ignorability} and \emph{Positivity}\cite{d2021overlap}\cite{pearl2009causal,rubin1974estimating}. With these four assumptions satisfied, we can successfully estimate the counterfactual outcome required by ITE.

	\section{Methodology}
	\label{Methods}
	\subsection{Overview of the Proposed CETransformer Model}
\begin{figure}[htb]
		\centering\includegraphics[width=\textwidth]{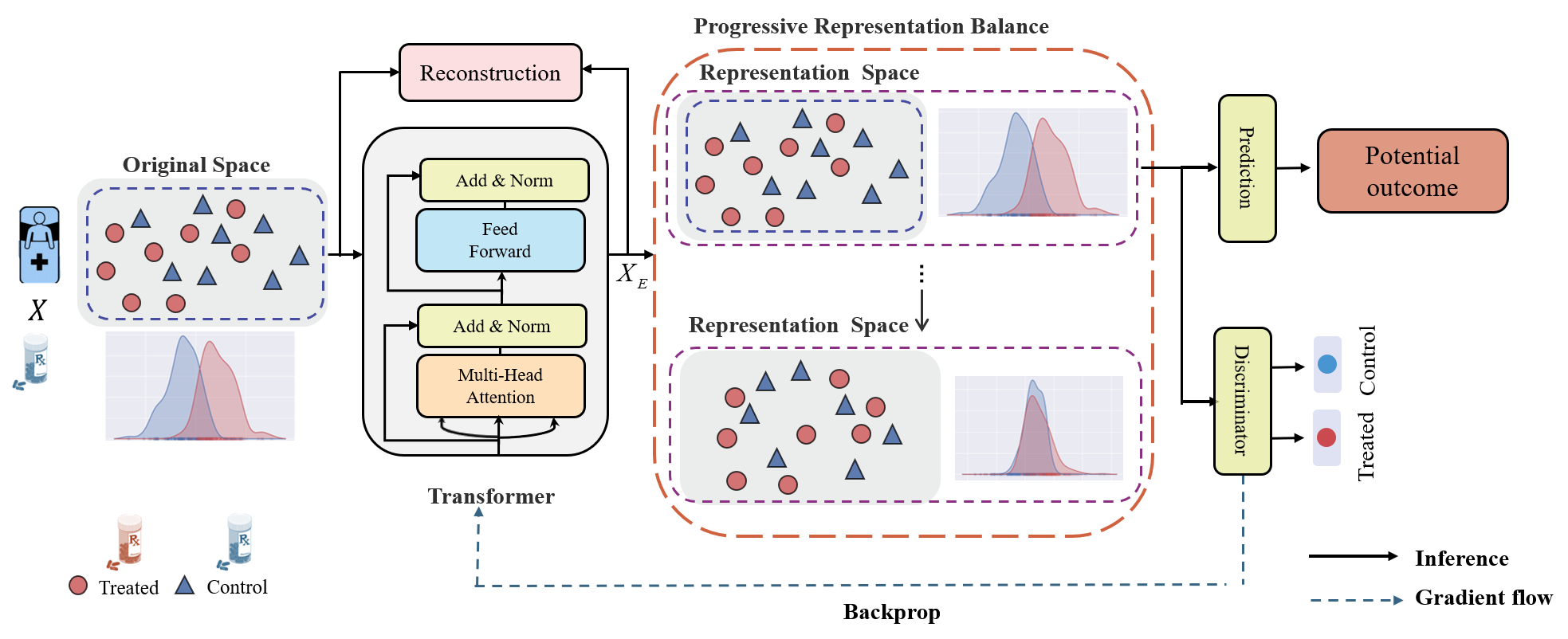}
		\caption{Illustration of the proposed CETransformer model for casual effect estimation via transformer based representation learning.}
		\label{fig:framework}
	\end{figure}

	In \cite{alaa2018limits}, it has been shown that the bound for the error in the estimation of individual causal effects mainly consists of the difference between the treated and control groups and the loss of outcome prediction. 
	Out of consideration of reducing the the difference between treated and control groups for robust estimation of counterfactual, we propose a CETransformer model for casual effect estimation via transformer based representation learning. As shown in Figure \ref{fig:framework}, the proposed framework of CETransformer contains three modules: 1) Self-supervised Transformer for representation learning which learns the balanced representation; 2) Discriminator network for adversarial learning to progressively shrink the difference between treated and control groups in the representation space; 3) Outcome prediction that uses the learned representations to estimate all potential outcome representations. For the details about each module, they will be presented in the following sections.
	
	\subsection{Self-supervised Transformer}
	
	Existing works, such as\cite{yao2018representation,yao2019ace}, learn representations of individuals through fully connected neural networks, and their core spirit is to balance the distribution between different treatment groups by means of carefully designed metric functions. Meanwhile, more attention should be paid to the correlation both between covariates and between individuals to generate more discriminative representation.  However, according to the theoretical analysis in\cite{domingos2020every}, simple fully connected networks only approximate learning the similarity function between samples. Based on the above 
	observations, we propose a  CETransformer model that is built upon transformers\cite{vaswani2017attention} to learn robust and balanced feature-contextual representation of individual features. Specifically, CETransformer models the corresponding correlations between different individuals features and obtains a robust representation of the individual by means of self-attention mechanism. 
	\begin{figure}[htb]
		\centering\includegraphics[width=0.8\textwidth]{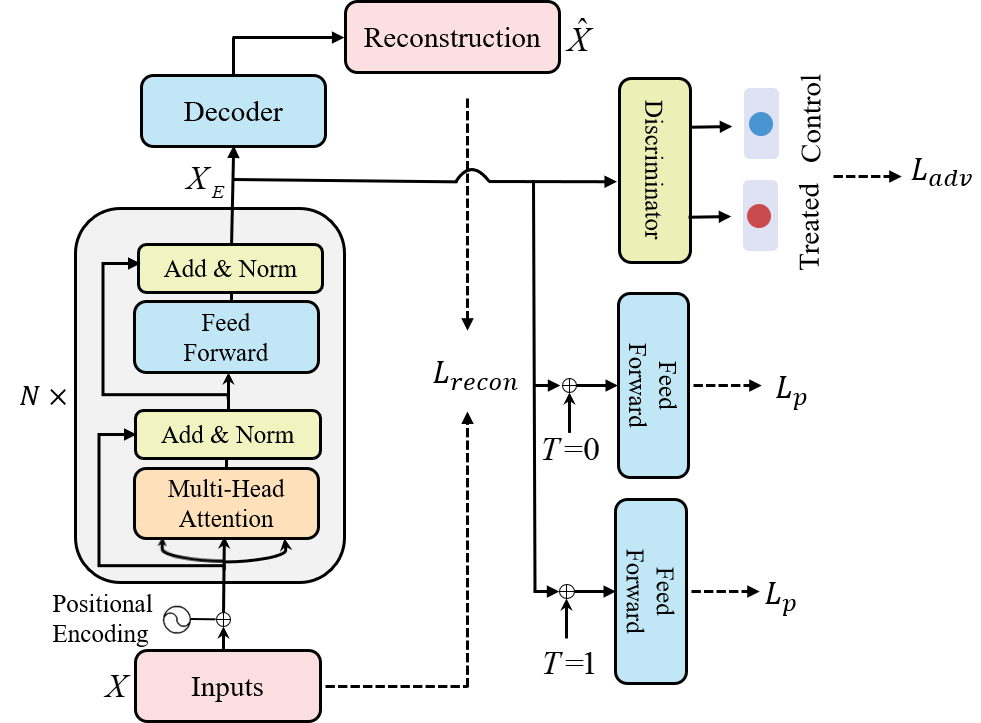}
		\caption{The architecture of the proposed adversarial transformer model.}
		\label{fig:network}
	\end{figure}
	
	As shown in Figure \ref{fig:network}, the CETransformer architecture comprises a stack of $N$ transformer blocks, a reconstruction feed-forward network, a discriminator network for adversarial learning, and an outcome prediction feed-forward network.  CETransformer first learns the embedding representation via the transformer $X_E = f_{trans}(X;\Theta_{trans})$, where $X_E$ denotes the embedding representation, and $f_{trans}(\cdot;\Theta_{trans})$ denotes the transformer with $\Theta_{trans}$ as its parameters. For a transformer, its core capability is to capture arbitrary distance dependencies through a self-attention mechanism:
	\begin{equation}
		Attention(Q,K,V) = softmax(\frac{Q\cdot K^T}{\sqrt{d_k}})V
	\end{equation}
	where $Q$, $K$, and $V$ represent Query, Key, Value, respectively, and $d_k$ stands for the dimension of Key. For $Q$, $K$, and $V$, all of them are obtained from the input $X$ through three different mapping networks.

	Since the training of transformer requires a large amount of data, to learn a robust representation for the transformer in the case of limited number of individuals is very difficult and even untrainable. In view of this situation,  the way of self-supervision is explored in the transformer framework to overcome this limitation to some extent. In particular, we adopt a simple fully connected network $f_{recon}(\cdot;\Theta_{dec})$ as a decoder to obtain a reconstructed  representation $\hat{X}=f_{recon}(X_E;\Theta_{dec})$, where $\Theta_{dec}$ is the network parameters. Here, MSE is used as the loss function to measure the reconstruction error:
	\begin{equation}
		L_{reco} = \left \| X - \hat{X} \right \|_{F}^2
	\end{equation}

	Compared with existing fully connected network-based approaches, our proposed CETransformer has the following advantages: 1) with the help of the self-attentive mechanism in Transformer, the correlations among different covariates are well exploited; 2) 
	by the way of self-supervised learning via auto-encoder to realize the augmentation of training data, the transformer can be trainable in the situation of limited number of individuals.

	\subsection{Adversarial Learning for Distribution Balancing}
	In order to satisfy the theoretical analysis mentioned in \cite{alaa2018limits} that the distribution of treated and control groups should overlap, the learned representations of the two groups should be balanced. Briefly, if it is not, there is a problem of covariates shift, which will lead to inaccurate estimation of potential outcomes. To the best of our knowledge, all existing works on causal effect estimation adopt a hand-crafted metric function to balance the distribution between the two groups. However, these approaches heavily rely on a 
	carefully designed metrics. Therefore, it is not a trivial task to deal with the complicated original distribution.
	
	Unlike the previous methods, we take adversarial learning to balance the distribution between the control and treated groups. Generative adversarial networks(GAN) is generally used to approximate the output distribution from the generator to the distribution of the real data. However, as far as causal reasoning is concerned, there is no such thing as real and generated data. To solve this problem, a straightforward way is to take the representation of the treated group as real data and the representation of the control group as generated data.

	For this case, to train a generative adversarial network, let $D(X_E^i)$ denote the discriminator network that maps the embedding representations of treated and control groups to the corresponding treatment assignment variables $T_i$. The discriminator network consists of a fully connected network, and the generator $G(\cdot)$ is the aforementioned transformer model. Due to the contradictory problems of the objective function in original GAN, which can lead to training instability and mode collapse. Many works\cite{arjovsky2017wasserstein,gulrajani2017improved,salimans2016improved} tries to solve these problems and in this paper we adopt WGAN\cite{arjovsky2017wasserstein} as our framework. Technically, WGAN minimizes a reasonable and efficient approximation of the Earth Mover(EM) distance, which is benefit for the stability of training. The loss function adopted by WGAN:
	\begin{equation}
		L_{adv} = \mathop{\mathrm{min}}\limits_{G} \,\mathop{\mathrm{max}}\limits_{D} \, \mathbb{E}_{X_E\sim \mathbb{P}_t}[D(X_E)] - \mathbb{E}_{\tilde{X}_E\sim \mathbb{P}_c}[D(\tilde{X_E})]
	\end{equation}
	where $\mathbb{P}_t$ and $\mathbb{P}_c$ represent the distributions of the treated and control groups, respectively. 
	
	\subsection{Outcome Prediction}
	After obtaining the balanced representation, we employed a two-branch network to predict the potential output $Y_i$ after a given $T_i$ based on the representation $X_E^i$ of the input $X_i$. Each branch is implemented by fully connected layers and one output regression layer.  Let $\tilde{Y_i} = h(X_E^i,T_i)$ denote the corresponding output prediction network. We aim to minimize the mean squared error in predicting factual outcomes:
	\begin{equation}
		L_p = \frac{1}{n} \sum_{i=1}^{n}(\tilde{Y_i} - Y_i)^2
	\end{equation}
	Ultimately, our total objective function can be expressed in the following form:
	\begin{equation}
		L = \alpha L_{reco} + \beta L_{adv} + \gamma L_p
	\end{equation}
	where the hyper-parameter $\alpha, \beta, \gamma$ controls the trade-off between the three function.
	
	\section{Experiments}
	\label{Experiments}
	\subsection{Datasets \& Metric}
	In this section, we conduct experiments on three public datasets which is same as \cite{zhang2020learning}, including the IHDP, Jobs, and Twins. On IHDP and Twins datasets, we average over 10 realizations with 61/27/10 ratio of train/validation/test splits. And on Jobs dataset, because of the extremely low treated/control ratio, we conduct the experiment on 10 train/validation/test splits with 56/24/20 split ratio, as suggested in \cite{shalit2017estimating}.
	
	The expected Precision in Estimation of Heterogeneous Effect(PEHE) \cite{hill2011bayesian} is adopted on IHDP and Twins dataset. The lower the $\varepsilon_{PEHE}$ is, the better the method is. On Jobs dataset, only the observed outcomes are available and the ground truth of ITE is unavailable. We adopt the policy risk \cite{shalit2017estimating} to measure the expected loss when taking the treatment as the ITE estimator suggests. Policy risk reflects how good the ITE estimation can guide the decision. The lower the policy risk is, the better the ITE estimation model can support the decision making.
	
	\begin{table}[htb]
		\caption{Mean performance (lower better) of individualized treatment effect estimation and standard deviation.}
		\label{results}
		\centering
		\resizebox{\textwidth}{!}{
			\begin{tabular}{c|cc|cccc}
				\hline
				& \multicolumn{2}{c|}{IHDP($\sqrt{\varepsilon_{PEHE}}$)}                     & \multicolumn{2}{c}{Twins($\sqrt{\varepsilon_{PEHE}}$)}                                                & \multicolumn{2}{c}{Jobs($\mathcal{R}_{pol}(\pi_f)$)}                        \\ \hline
				& In-sample              & Out-sample             & In-sample                & \multicolumn{1}{c|}{Out-sample}               & In-sample              & Out-sample             \\ \hline
				OLS/LR$_1$           & 5.8 $\pm$ .3           & 5.8 $\pm$ .3           & .319 $\pm$ .001          & \multicolumn{1}{c|}{.318 $\pm$ .007}         & .22 $\pm$ .00          & .23 $\pm$ .02          \\
				OLS/LR$_2$           & 2.4 $\pm$ .1           & 2.5 $\pm$ .1           & .320 $\pm$ .002          & \multicolumn{1}{c|}{.320 $\pm$ .003}          & .21 $\pm$ .00          & .24 $\pm$ .01          \\
				BLR                 & 5.8 $\pm$ .3           & 5.8 $\pm$ .3           & .312 $\pm$ .003          & \multicolumn{1}{c|}{.323 $\pm$ .018}          & .22 $\pm$ .01          & .26 $\pm$ .02          \\
				K-NN                & 2.1 $\pm$ .1           & 4.1 $\pm$ .2           & .333 $\pm$ .001          & \multicolumn{1}{c|}{.345 $\pm$ .007}          & .22 $\pm$ .00          & .26 $\pm$ .02          \\
				BART                & 2.1 $\pm$ .1           & 2.3 $\pm$ .1           & .347 $\pm$ .009          & \multicolumn{1}{c|}{.338 $\pm$ .016}          & .23 $\pm$ .00          & .25 $\pm$ .02          \\
				R-FOREST            & 4.2 $\pm$ .2           & 6.6 $\pm$ .3           & .366 $\pm$ .002          & \multicolumn{1}{c|}{.321 $\pm$ .005}          & .23 $\pm$ .01          & .28 $\pm$ .02          \\
				C-FOREST            & 3.8 $\pm$ .2           & 3.8 $\pm$ .2           & .366 $\pm$ .003          & \multicolumn{1}{c|}{.316 $\pm$ .011}          & .19 $\pm$ .00          & .20 $\pm$ .02          \\
				TARNET              & .88 $\pm$ .02          & .95 $\pm$ .02          & .317 $\pm$ .002          & \multicolumn{1}{c|}{.315 $\pm$ .003}          & .17 $\pm$ .01          & .21 $\pm$ .01          \\
				CAR$_{WASS}$           & .72 $\pm$ .02          & .76 $\pm$ .02          & .315 $\pm$ .007          & \multicolumn{1}{c|}{.313 $\pm$ .008}          & .17 $\pm$ .01          & .21 $\pm$ .01          \\
				SITE                & .60 $\pm$ .09          & .65 $\pm$ .10          & .309 $\pm$ .002          & \multicolumn{1}{c|}{.311 $\pm$ .004}          & .22 $\pm$ .00          & .22 $\pm$ .00          \\
				ACE                 & .49 $\pm$ .04          & .54 $\pm$ .06          & .306 $\pm$ .000          & \multicolumn{1}{c|}{.301 $\pm$ .002}          & .21 $\pm$ .00          & .21 $\pm$ .00          \\
				DKLITE              & .52 $\pm$ .02          & .65 $\pm$ .03          & .288 $\pm$ .001          & \multicolumn{1}{c|}{.293 $\pm$ .003}          & .13 $\pm$ .01          & .14 $\pm$ .01          \\ \hline
				CETransformer(Ours) & \textbf{.46 $\pm$ .02} & \textbf{.51 $\pm$ .03} & \textbf{.287 $\pm$ .001} & \multicolumn{1}{c|}{\textbf{.289 $\pm$ .002}} & \textbf{.12 $\pm$ .01} & \textbf{.13 $\pm$ .00} \\ \hline
			\end{tabular}
		}
	\end{table}

	\subsection{Competing Algorithms}
	We compare CETransformer with a total of 12 algorithms. First we evaluate least squares regression using treatment as an additional input feature (OLS/LR$_1$), we consider separate least squares regressions for each treatment (OLS/LR$_2$), we evaluate balancing linear regression (BLR)\cite{johansson2016learning}, k-nearest neighbor (k-NN) \cite{crump2008nonparametric}, Bayesian additive regression trees (BART)\cite{chipman2010bart}, random forests (R-Forest)\cite{breiman2001random}, causal forests (C-Forest)\cite{wager2018estimation}, treatment-agnostic representation network (TARNET), counterfactual regression with Wasserstein distance(CARW ASS)\cite{shalit2017estimating}, local similarity preserved individual treatment effect (SITE)\cite{yao2018representation}, adaptively similarity-preserved representation learning method for Causal Effect estimation (ACE)\cite{yao2019ace}, deep kernel learning for individualized treatment effects(DKLITE)\cite{zhang2020learning}.
	
	\begin{figure}[htb]
		\centering\includegraphics[width=0.8\textwidth]{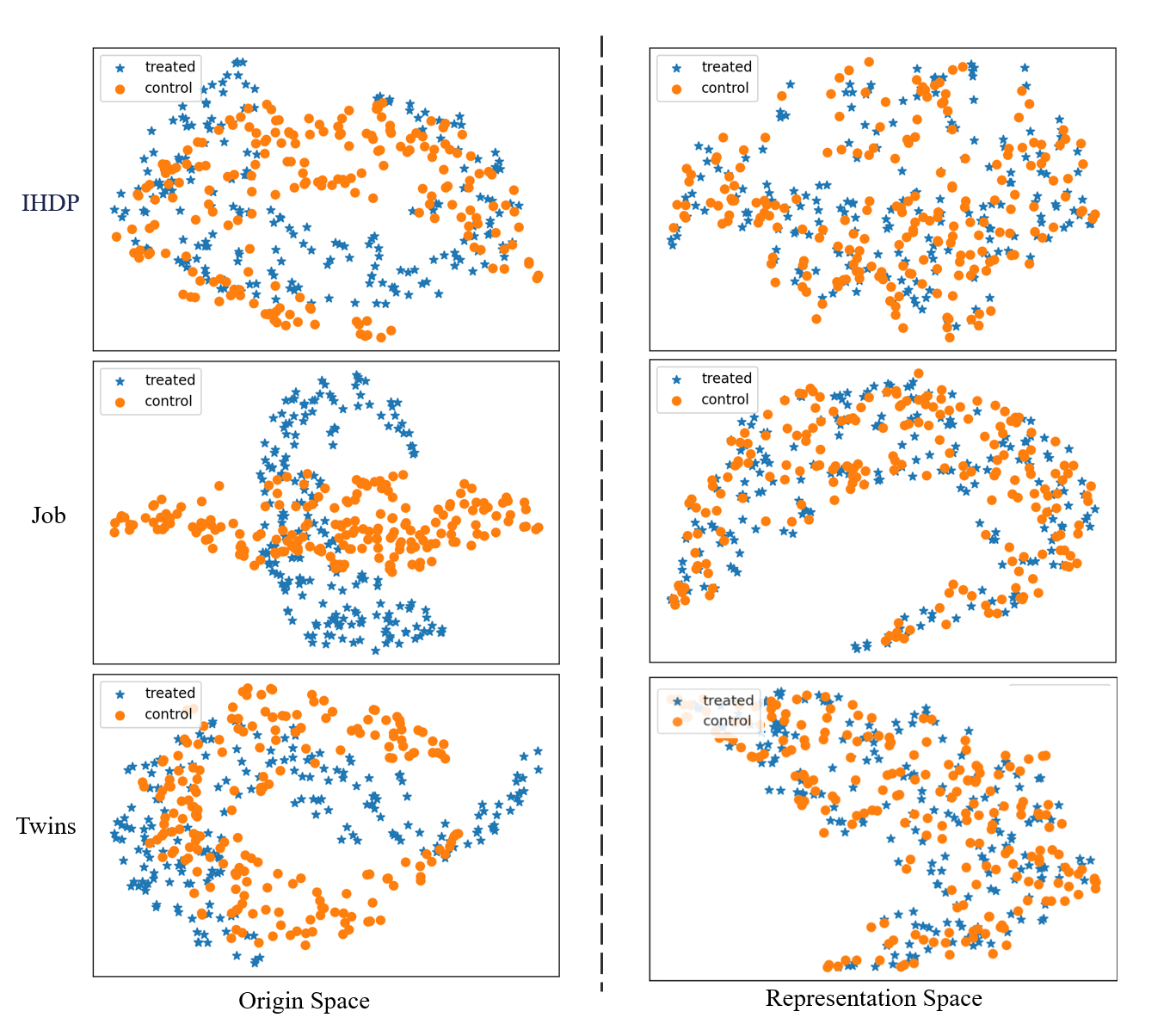}
		\caption{\textbf{Visualization of the learned representations in the three datasets through t-SNE.} By sampling the dataset and performing dimensionality reduction using t-SNE, we can find that the representations learned by CEtransformer for different treatment groups are highly overlapping, which is in accordance with our expectation.}
		\label{fig:representation}
	\end{figure}
	
	\subsection{Prediction Performance Results}
	With the same settings as \cite{zhang2020learning}, we report in-sample and out-of-sample performance in Table \ref{results}. CETransformer adopts the transformer network as a backbone and learns balanced representations via adversarial learning. With this approach, we outperform all competing algorithms on each benchmark dataset. Probably the most relevant of these comparisons are the three works\cite{yao2018representation,yao2019ace,zhang2020learning}, which generate overlapping representations of the treatment and control groups by means of neural networks and hand-designed inter-distributional similarity metric functions. Compared to them, the improvement in performance better highlights the predictive power of our representation. In addition to performance, we are also interested in whether the learned representations are balanced. Figure \ref{fig:representation} shows the visualization of the learned representations in the three datasets through t-SNE\cite{van2008visualizing}. Stars and circles represent the two-dimensional representations of the treated and control groups respectively, and we can find that the distance between the two distributions is well approximated by the adversarial learning of CETransformer, which indicates that no covariate shift occurs. 
	
	\begin{table}[htb]
		\caption{Ablation on CETransformer: Performance Comparison on Three Datasets.}
		\label{Ablation}
		\centering
		
		\begin{tabular}{c|c|c|c|c}
			\hline
			Dataset                &            & CETransformer     & \begin{tabular}[c]{@{}c@{}}Without \\ Transformer\end{tabular} & \begin{tabular}[c]{@{}c@{}}Without \\ Discriminator\end{tabular} \\ \hline
			\multirow{2}{*}{IHDP}  & In-sample  & .46 $\pm$ .02   & .50 $\pm$ .03                                                  & 2.8 $\pm$ .13                                                    \\
			& Out-sample & .51 $\pm$ .03   & .56 $\pm$ .03                                                  & 2.9 $\pm$ .22                                                    \\ \hline
			\multirow{2}{*}{Twins} & In-sample  & .287 $\pm$ .001 & .292 $\pm$ .001                                                & .335 $\pm$ .003                                                  \\
			& Out-sample & .289 $\pm$ .002 & .295 $\pm$ .002                                                & .317 $\pm$ .012                                                  \\ \hline
			\multirow{2}{*}{Jobs}  & In-sample  & .12 $\pm$ .01   & .13 $\pm$ .01                                                  & .18 $\pm$ .00                                                    \\
			& Out-sample & .13 $\pm$ .00   & .15 $\pm$ .01                                                  & .20 $\pm$ .02                                                    \\ \hline
		\end{tabular}
		
	\end{table}
	
	\begin{figure}[htb]
		\centering\includegraphics[width=0.8\textwidth]{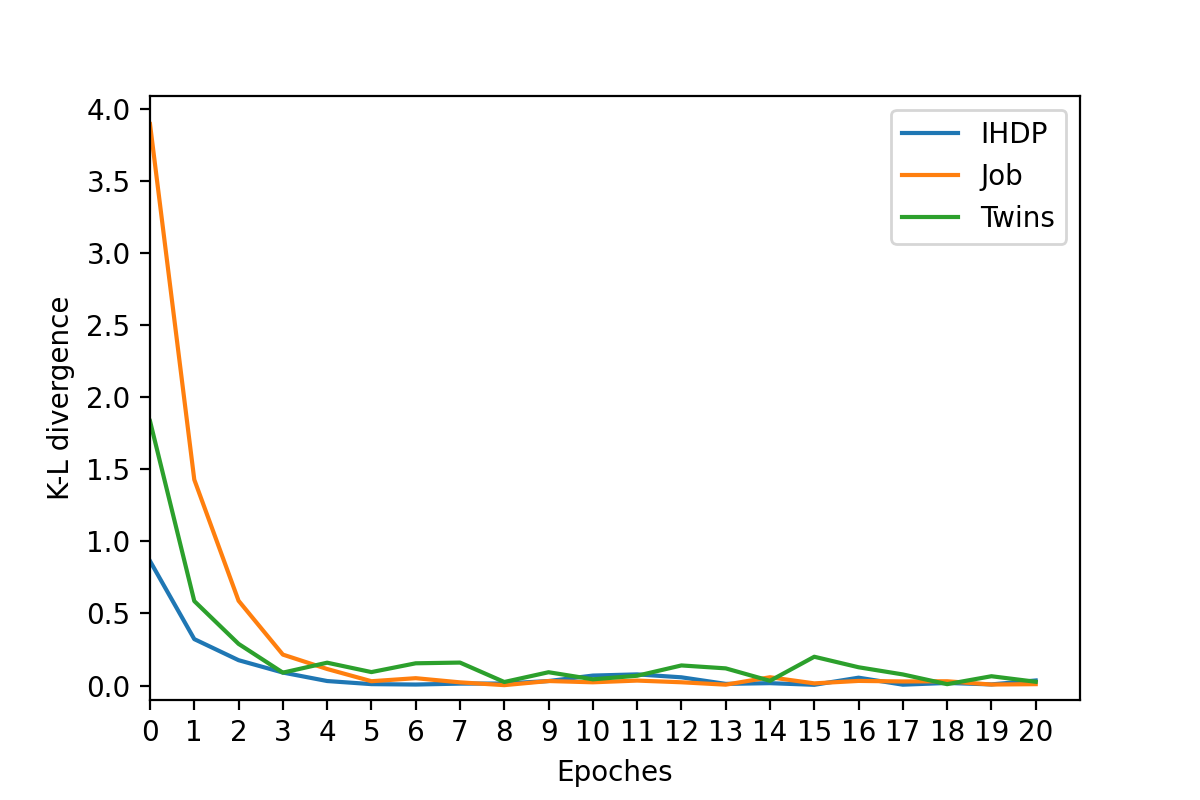}
		\caption{\textbf{KL divergence between control and treated groups in representation space.} }
		\label{fig:KL}
	\end{figure}
	
	Meanwhile, quantitative metric, i.e., K-L divergence,  of the different treatment groups is given in Figure \ref{fig:KL}. We can find that the difference  between the treated and control groups in the representation space decreases rapidly with the number of training iterations. As we inferred from the t-SNE visualization results, the differences in the distributions between the different treatment groups are wiped out under the constraints of the adversarial network.

	\subsection{Ablation Study}
	Experimental results on three datasets show that CETransformer is able to estimate causal effects more accurately compared to existing representation-based learning methods. In this section, we further explore the extent to which the changes made in CETransformer: Transformer backbone and adversarial learning affect the results. We compare CETransformer with CETransformer without Transformer and CETransformer without adversarial learning. Table \ref{Ablation} shows the results.

	Our motivation for replacing the fully connected network with Transformer is that Transformer's self-attention can better capture the correlation between different features, and that correlation is expected to yield a more predictive representation. The results confirm our conjecture, and we find that replacing the transformer with a fully connected network will cause a different degree of performance degradation.  Then, the distribution of the two groups is balanced using an adversarial learning  rather than a hand-crafted metric function. When the adversarial learning part is removed, the distribution imbalance should exist in the representation space as shown by the theoretical analysis in \cite{alaa2018limits}. The experimental results also confirm our conjecture, and it can be found that after removing the adversarial learning part, the performance of CETransfoermer is similar to that of using traditional supervised learning methods alone.

	\section{Conclusion}
	\label{Conclusion}
	In many domains, understanding the effect of different treatments on the individual level is crucial, but predicting their potential outcome in real-life is challenging. In this paper, we propose a novel balanced representation distribution learning model based on Transformer for the estimation of individual causal effects. We fully exploit the correlation information among the input features by the Transformer structure and automatically balance the distribution between the treated and control  groups by the adversarial learning. Extensive experiments on three benchmark datasets show that CETransformer outperforms the state-of-the-art methods, which demonstrates the competitive level of CETransformer in estimating causal effects. We further demonstrate the effectiveness of components in CEtransformer through ablation study.

	\bibliographystyle{splncs04}
	\bibliography{mybib.bib}
\end{document}